\pdfoutput=1

\documentclass[11pt]{article}

\usepackage[final]{ACL2023}

\usepackage{times}
\usepackage{latexsym}
\usepackage{diagbox}

\usepackage[utf8]{inputenc}
\usepackage{graphicx}
\usepackage{pgfplots}
\usepackage{booktabs}
\usepackage{float}
\usepackage{listings}
\usepackage{tabularx}
\usepackage{booktabs} 
\usepackage{array} 

\usepackage{tikz}
\usepackage{amsmath}
\usepackage{graphicx}
\usepackage{adjustbox}

\usepackage[T1]{fontenc}

\usepackage[utf8]{inputenc}

\usepackage{microtype}

\usepackage{url}            
\usepackage{microtype}
\usepackage{amsmath}
\usepackage{amssymb}
\usepackage {tcolorbox}
\tcbuselibrary{theorems}
\usepackage{graphicx}
\usepackage{comment}
\usepackage{array}
\usepackage{mathtools}
\usepackage{csquotes}
\usepackage{multirow}
\usepackage{enumitem}
\usepackage{amsmath}
\usepackage{tcolorbox}
\usepackage{xcolor}

\title{RARE: Retrieval-Augmented Reasoning Enhancement for Large Language Models}

\author{
Hieu Tran$^1$, Zonghai Yao$^1$, Zhichao Yang$^1$, Junda Wang$^1$,
Yifan Zhang$^2$\\ \textbf{Shuo Han}$^2$, \textbf{Feiyun Ouyang}$^2$, \textbf{Hong Yu$^{1,2}$}\\
$^1$University of Massachusetts, Amherst\\
    $^2$University of Massachusetts, Lowell\\
}

\begin{document}
\maketitle
\begin{abstract}
This work introduces \textbf{RARE (Retrieval-Augmented Reasoning Enhancement)}, a versatile extension to the mutual reasoning framework (rStar), aimed at enhancing reasoning accuracy and factual integrity across large language models (LLMs) for complex, knowledge-intensive tasks such as medical and commonsense reasoning. RARE incorporates two innovative actions within the Monte Carlo Tree Search 
framework: \textbf{(A6)}, which generates search queries based on the initial problem statement, performs information retrieval using those queries, and augments reasoning with the retrieved data to formulate the final answer; and \textbf{(A7)}, which leverages information retrieval specifically for generated sub-questions and re-answers these sub-questions with the relevant contextual information. Additionally, a Retrieval-Augmented Factuality Scorer is proposed to replace the original discriminator, prioritizing reasoning paths that meet high standards of factuality. Experimental results with LLaMA 3.1 show that RARE enables open-source LLMs to achieve competitive performance with top closed-source models like GPT-4 and GPT-4o. This research establishes RARE as a scalable solution for improving LLMs in domains where logical coherence and factual integrity are critical~\footnote{Our code will be released at: \url{https://github.com/fatebreaker/RARE}}.
\end{abstract}

\section{Introduction}
\begin{figure*}
    \centering
    \includegraphics[width=\linewidth]{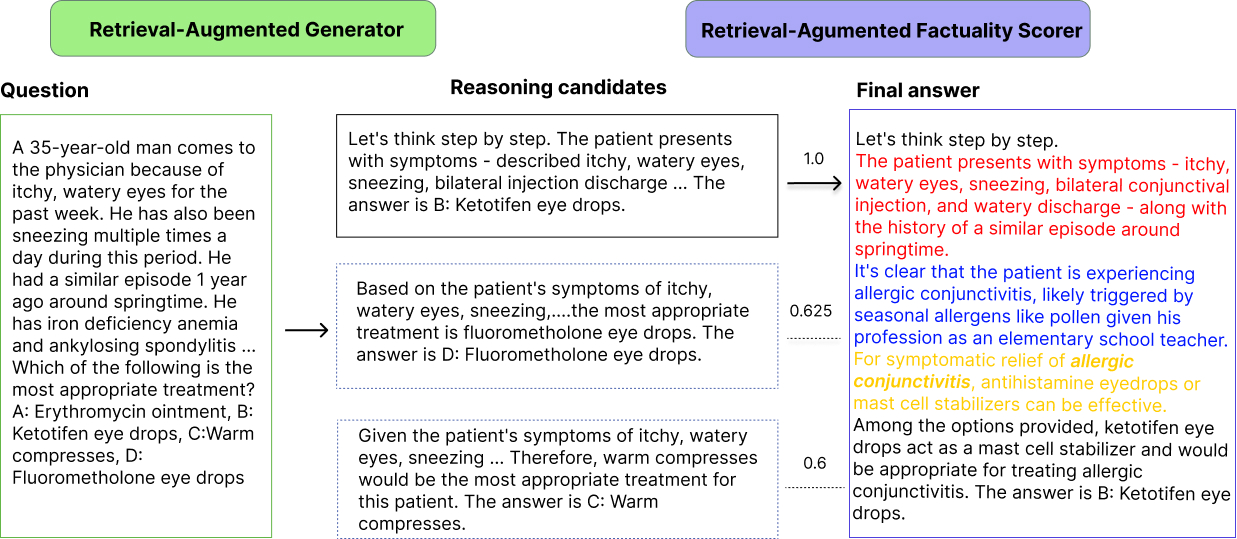}
    \caption{Overview of our reasoning process, which combines generation and factuality scoring. (1) A retrieval-augmented generator produces multiple candidate reasoning paths using Monte Carlo Tree Search (MCTS); (2) a retrieval-augmented factuality scorer evaluates the factual accuracy of each reasoning path; (3) the trajectory with the highest factuality score is selected as the final answer.}
    \label{fig:overview}
\end{figure*}

Question answering (QA) is a cornerstone task in natural language processing that involves generating answers to questions posed in natural language. QA spans a broad spectrum of domains and types, ranging from open-domain QA \cite{yang2018hotpotqa, kwiatkowski2019natural}
to more specialized areas like medical QA \cite{jin2021disease, cao2011askhermes}. The overwhelming volume and complexity of medical information necessitate medical QA, which benefits many downstream tasks such as medical education, clinical decision support, and patient care optimization \cite{cai2023paniniqa, liu2023using, jin2024matching}.

Medical QA represents a unique and demanding subset of QA, requiring models to navigate intricate medical knowledge, interpret clinical scenarios, and select correct and contextually appropriate options \cite{singhal2023towards, wu2024pmc}. Similar to general domain QA, Medical QA requires structured multi-step reasoning, where answers emerge from various sequential steps. Take Figure ~\ref{fig:overview} as an example, to find appropriate treatment given patient information, the QA model should first identify patient conditions (colored in red, e.g., chief complaint and past conditions), then analyze contributing factors and diagnose the disease (colored in blue), and determine appropriate evidence-based interventions in the final step (colored in yellow). Without such structured multi-step reasoning, it would be challenging to arrive at an accurate and contextually relevant answer for such a complex medical question.

Moreover, Medical QA presents two non-trivial challenges that distinguish it from general-domain QA. First, Medical QA depends heavily on domain-specific knowledge that is not always available within pre-trained models, necessitating knowledge-based retrieval from external sources \cite{xiong2024benchmarking}. Figure ~\ref{fig:overview} is an example which involves specific medical terms such as \textit{allergic conjunctivitis}. In addition, medical knowledge evolves rapidly, and new treatments or updated guidelines may not be included in the model's pretrained corpus. For example, newer drugs (like epinastine hydrochloride for allergic conjunctivitis) may be recommended by recent guidelines but absent in older pre-trained models. Second, Medical QA encompasses a wide variety of question types, including not only multi-step reasoning and  knowledge-based retrieval as previously mentioned, but also composite questions requiring iterative evidence retrieval, where they demand retrieval of relevant knowledge at each reasoning step to ensure accuracy and relevance throughout the process. 

In parallel, Commonsense Question Answering shares similar complexities with Medical QA, particularly in its reliance on structured multi-step reasoning and iterative evidence retrieval. 
While Medical QA draws heavily on domain-specific knowledge, Commonsense QA focuses on leveraging a model's understanding of general world knowledge and logical connections to answer questions that are often indirect or abstract. For example, tasks like StrategyQA \cite{geva2021did} require models to infer hidden relationships and execute multi-hop reasoning, akin to diagnosing a condition in Medical QA \cite{trivedi2023interleaving, bauer2018commonsense, chen2020improving}. 
This similarity in reasoning processes across both domains underscores the importance of designing frameworks that can adapt to and optimize multi-step reasoning workflows, irrespective of the domain.


\begin{figure*}
    \centering
    \includegraphics[width=1.0\linewidth]{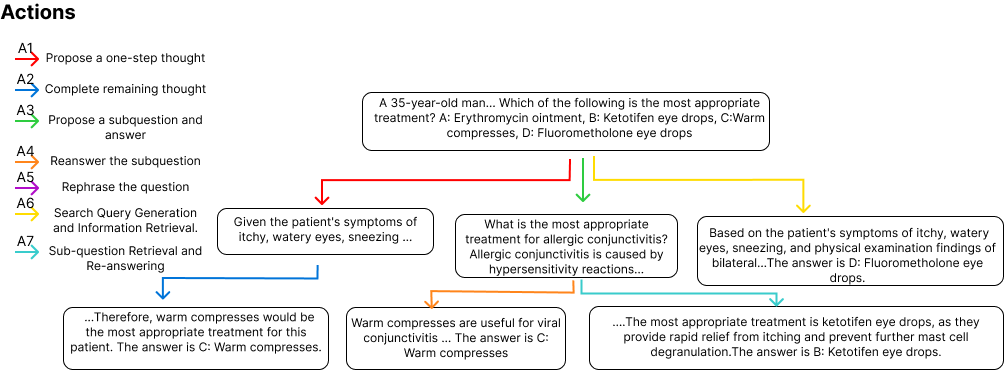}
    \caption{An example to illustrate the process of retrieval-augmeted generator. Highlighted nodes from top to bottom
constitute a complete reasoning trace. Given a question, MCTS augments the LLM to explore a rich, human-like reasoning action space and generate the next steps based on the current state.}
    \label{fig:generator}
\end{figure*}

In this paper, we propose Retrieval-Augmented Reasoning Enhancement (RARE) to handle aforementioned challenges. We built upon rStar \cite{qi2024mutual} where a language model generates reasoning steps and another verifies them, improving accuracy without fine-tuning or superior models. To answer knowledge-based questions, RARE designed a new action A6, which generates multiple search queries based on the question and retrieves relevant documents. To answer composite questions, we add action A7, which refines sub-questions, retrieves targeted information, and updates the next step. RARE applies the Monte Carlo Tree Search (MCTS) algorithm to select the best action path that leads to the final answer. In addition, RARE is complemented by Retrieval-Augmented Factuality Scorer (RAFS) that evaluates and ranks reasoning paths for factual accuracy. 

We applied RARE and other baselines on 3 medical QA tasks and 3 general domain QA tasks. Results show that RARE significantly enhances accuracy across various LLMs, enabling the open-source LLMs (LLaMA 3) to achieve competitive performance with top closed-source LLMs like GPT-4o. Our contributions are as follows:

\begin{enumerate}
    \item \textbf{Formulating Medical QA as Multi-Step Reasoning}: We build upon the rStar framework to model medical QA as a structured multi-step reasoning task, addressing the complexity and sequential nature of medical queries. 
    \item \textbf{Novel Retrieval Actions}: We introduce two retrieval-augmented actions within the MCTS framework, enabling the integration of real-time, context-specific information to enhance reasoning accuracy and relevance. 
    \item \textbf{Retrieval-Augmented Factuality Scorer}: We propose a Retrieval-Augmented Factuality Scorer to evaluate and rank reasoning paths, ensuring they maintain both logical coherence and factual reliability throughout the reasoning process. 
\end{enumerate}


\section{Preliminaries}

This section introduces the foundational concepts and notations used in this work, focusing on the Monte Carlo Tree Search (MCTS) algorithm and the rStar framework \cite{qi2024mutual}, which serve as the basis for our proposed RARE method.
\subsection{Monte Carlo Tree Search (MCTS)}

Monte Carlo Tree Search (MCTS) is a decision-making algorithm that incrementally builds a search tree by simulating outcomes to estimate action values, making it effective for complex reasoning tasks \cite{browne2012survey}. It operates through iterative selection, expansion, simulation, and backpropagation phases, balancing exploration and exploitation via the Upper Confidence Bound applied on Trees (UCT). MCTS enables adaptive strategy refinement in large search spaces where direct computation is infeasible. A detailed explanation of MCTS and its implementation can be found in the Appendix ~\ref{subsec:a5}.

\subsection{Mutual Reasoning Makes Smaller LLMs Stronger Problem-Solvers}

Building upon MCTS, \cite{qi2024mutual} proposed rStar, a framework that augments MCTS with a diverse set of reasoning actions. This enhancement is designed to improve exploration of the solution space in complex reasoning tasks by allowing more dynamic and human-like reasoning pathways. Traditional approaches, such as Chain of Thought (CoT) reasoning \cite{wei2022chain} or self-consistency \cite{wang2022self}, often rely on a single action type, which can limit the diversity and effectiveness of generated solutions. In contrast, rStar incorporates five distinct actions that enable more adaptive exploration(the prompt for each action can be found in Appendix ~\ref{subsec:a7}): 

\begin{itemize}[leftmargin=.1in,topsep=0.3pt]
\setlength\itemsep{0em}
    \item \textbf{A1: Propose a One-Step Thought}. This action generates the next reasoning step based on previous steps, allowing the LLM to build the solution incrementally. 
    \item \textbf{A2: Propose Remaining Thought Steps}. This action enables the LLM to produce all remaining reasoning steps in one inference, similar to CoT, for simpler questions. 
    \item \textbf{A3: Generate Next Sub-question and Answer}. This action decomposes the main problem into a sequence of sub-questions, each solved in turn.
    \item \textbf{A4: Re-answer Sub-question}. This action allows the LLM to re-answer a previously generated sub-question, increasing accuracy by using few-shot prompting. 
    \item \textbf{A5: Rephrase Question/Sub-question}. This action rephrases the question to clarify conditions and reduce misunderstandings, enhancing the LLM's interpretation of the problem.
\end{itemize}

\begin{figure}[!ht]
    \centering
    \includegraphics[width=\linewidth]{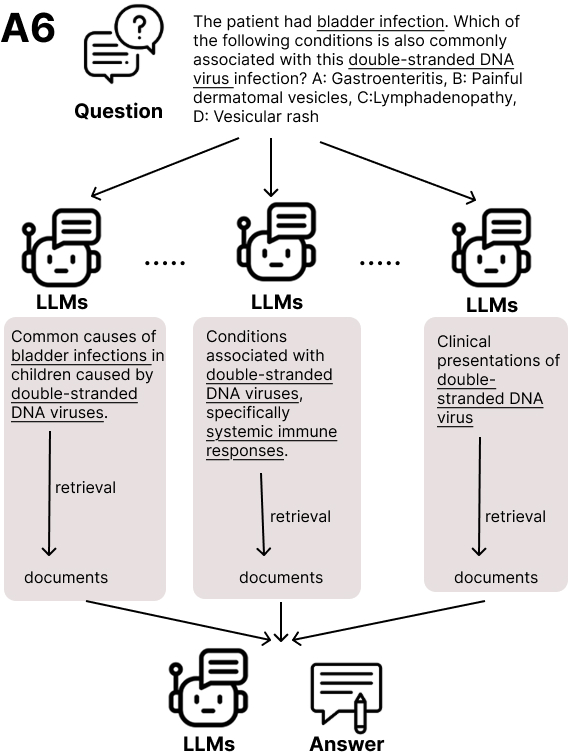}
    \caption{Illustration of the proposed retrieval-augmented action A6 in RARE: Given a question, LLMs generate search queries and retrieve relevant documents to construct a contextually enriched final answer. Key entities are underlined.}
    \label{fig:a6}
\end{figure}

\section{Methodology}

\subsection{Overview of RARE Framework}
Inspired by the generator-discriminator structure of rStar \cite{qi2024mutual}, RARE introduces a retrieval-augmented generator and a factuality scorer to enhance reasoning accuracy and factual integrity in large language models. As illustrated in Figure \ref{fig:overview}, RARE operates in two main stages: 

\noindent\textbf{Candidate Generation with Retrieval-Augmented Generator}: The retrieval-augmented generator builds on the MCTS-based rStar self-generator, incorporating two new retrieval-augmented actions that dynamically fetch relevant external information. These actions improve the relevance and accuracy of candidate reasoning paths by integrating contextually enriched knowledge into intermediate reasoning steps, especially for complex questions, illustrated in Figure \ref{fig:generator}. 
 
\paragraph{Factuality Evaluation with Retrieval-Augmented Factuality Scorer}: Replacing the discriminator in rStar, the Retrieval-Augmented Factuality Scorer evaluates each candidate trajectory's factual reliability. This scorer verifies the alignment of intermediate reasoning steps with retrieved evidence, assigning a factuality score that reflects the trajectory’s consistency with external knowledge. The trajectory with the highest factuality score is selected as the final answer, prioritizing the most factually supported reasoning path. This selection ensures coherence and factual alignment, enhancing response reliability.

Through these stages, RARE systematically integrates retrieval-based evidence into the reasoning process, optimizing both reasoning coherence and factual accuracy. This approach makes RARE well-suited for knowledge-intensive tasks, such as commonsense and medical reasoning.

\begin{figure}[!ht]
    \centering
    \includegraphics[width=\linewidth]{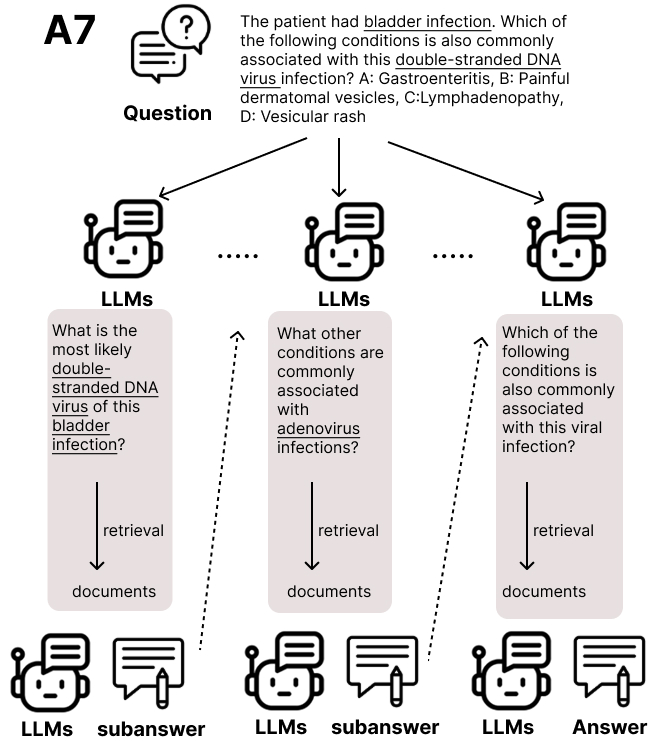}
    \caption{Illustration of the proposed retrieval-augmented action A7 in RARE: LLMs decompose the question into sub-questions, perform retrieval for each sub-question, and re-answer them based on the retrieved documents. The final sub-question is a rephrased version of the original question, so the sub-answer to this final sub-question also serves as the answer to the original question. In comparison with previous figure, we can find that A6 tends to use existing entity from the main question where A7 tends to use additional entity from previous subanswer.}
    \label{fig:a7}
\end{figure}

\begin{figure*}[h!]
\centering
\adjustbox{max width=\textwidth}{
\begin{tikzpicture}[node distance=2.0cm and 1.0cm] 

    \node[draw, rounded corners, fill=red!10, inner sep=10pt, text width=16cm, align=center] (RAFS) at (0, 5) {\textbf{Retrieval-Augmented Factuality Scorer (RAFS)}};

    \node[align=center, font=\bfseries, text width=3.8cm] (Step1) at (-5.5, 3.5) {1. Split into \\ statements};
    
    \node[draw, align=left, rounded corners, fill=white, text width=2.5cm, minimum height=2.0cm] (Fact1) at (-5.5, 2.0) {Given the patient's symptoms of itchy, watery eyes\dots};
    \node[draw, align=left, rounded corners, fill=white, text width=2.5cm, minimum height=2.0cm] (Fact2) at (-5.5, -0.3) {The best treatment for mild allergic conjunctivitis \dots};
    \node[draw, align=left, rounded corners, fill=white, text width=2.5cm, minimum height=2.0cm] (Fact3) at (-5.5, -2.6) {Warm compresses are often recommended \dots};
    \node[draw, align=left, rounded corners, fill=white, text width=2.5cm, minimum height=2.0cm] (Fact4) at (-5.5, -4.9) {Therefore, warm compresses would be the most\dots};
    \node[draw, align=left, rounded corners, fill=white, text width=2.5cm, minimum height=2.0cm] (Fact5) at (-5.5, -7.2) {The answer is C: Warm compresses.};

    \node[align=center, font=\bfseries, text width=3.8cm] (Step2) at (-2.0, 3.5) {2. Generate retrieval queries};

    \node[draw, align=center, rounded corners, fill=white, text width=3.5cm, minimum height=2.0cm] (Revise1) at (-2.0, 2.0) {Treatment options for seasonal\dots\\ Seasonal allergic conjunctivitis\dots};
    \node[draw, align=center, rounded corners, fill=white, text width=3.5cm, minimum height=2.0cm] (Revise2) at (-2.0, -0.3) {Best treatment for mild allergic\dots\\ \dots avoiding triggers lubricating artificial};
    \node[draw, align=center, rounded corners, fill=white, text width=3.5cm, minimum height=2.0cm] (Revise3) at (-2.0, -2.6) {Warm compresses treatment allergic conjunctivitis effectiveness\dots};
    \node[draw, align=center, rounded corners, fill=white, text width=3.5cm, minimum height=2.0cm] (Revise4) at (-2.0, -4.9) {Treatment options for allergic conjunctivitis\dots};
    \node[draw, align=center, rounded corners, fill=white, text width=3.5cm, minimum height=2.0cm] (Revise5) at (-2.0, -7.2) {Most effective treatments for allergic conjunctivitis warm compresses\dots};

    \node[align=center, font=\bfseries, text width=2.8cm] (Step3) at (1.6, 3.5) {3. Retrieve information};

    \node[draw, align=center, rounded corners, fill=white, text width=3.0cm, minimum height=2.0cm] (Relevance1) at (1.8, 2.0) {\dots control symptoms such as sneezing, itching \dots};
    \node[draw, align=center, rounded corners, fill=white, text width=3.0cm, minimum height=2.0cm] (Relevance2) at (1.8, -0.3) {\dots nedocromil as mast cell stabilizers, which come as eye drops.};
    \node[draw, align=center, rounded corners, fill=white, text width=3.0cm, minimum height=2.0cm] (Relevance3) at (1.8, -2.6) {If the allergen is encountered and the symptoms are mild \dots
};
    \node[draw, align=center, rounded corners, fill=white, text width=3.0cm, minimum height=2.0cm] (Relevance4) at (1.8, -4.9) {\dots allergic conjunctivitis may also require topical steroid drops.
};
     \node[draw, align=center, rounded corners, fill=white, text width=3.0cm, minimum height=2.0cm] (Relevance5) at (1.8, -7.2) {\dots stabilizers can help people with allergic conjunctivitis \dots};

    \node[align=center, font=\bfseries, text width=3.8cm] (Step4) at (5.5, 3.5) {4. Rate using retrieved information};

    \node[draw, align=left, rounded corners, fill=green!20, text width=3.0cm, minimum height=2.0cm] (Rate1) at (5.5, 2.0) {Given the patient's symptoms of itchy, watery eyes \dots};
    \node[draw, align=left, rounded corners, fill=green!20, text width=3.0cm, minimum height=2.0cm] (Rate2) at (5.5, -0.3) {The best treatment for mild allergic conjunctivitis \dots};
    \node[draw, align=left, rounded corners, fill=green!20, text width=3.0cm, minimum height=2.0cm] (Rate3) at (5.5, -2.6) {Warm compresses are often recommended \dots};
    \node[draw, align=center, rounded corners, fill=red!20, text width=3.0cm, minimum height=2.0cm] (Rate4) at (5.5, -4.9) {Therefore, warm compresses would be the most \dots};
    \node[draw, align=center, rounded corners, fill=red!20, text width=3.0cm, minimum height=2.0cm] (Rate5) at (5.5, -7.2) {The answer is C: Warm compresses.};

    \node[draw, rounded corners, fill=blue!10, align=left, text width=3.0cm, minimum height=2.7cm] (Prompt) at (-9, 1.7) {\textbf{Question}\\ A 35-year-old man comes \dots Which of the following is the most appropriate treatment?};
    \node[draw, rounded corners, fill=blue!10, align=left, text width=3.0cm, minimum height=4.7cm] (Response) at (-9, -3.3) {\textbf{Reasoning}\\ Given the patient's symptoms of itchy, watery eyes, sneezing, and conjunctival \dots The answer is C: Warm compresses.};

    \node[draw, rounded corners, fill=blue!10, align=left, text width=3.4cm, minimum height=2.6cm] (Output) at (9.5, -1.0) {\textbf{Output}\\ Supported: 3 \\ Not Supported: 2 \\ Factuality Score: 0.6};

    \draw[-, thick] (Response.east) -- (Fact1.west);
    \draw[-, thick] (Response.east) -- (Fact2.west);
    \draw[-, thick] (Response.east) -- (Fact3.west);
    \draw[-, thick] (Response.east) -- (Fact4.west);
    \draw[-, thick] (Response.east) -- (Fact5.west);

    \draw[-, thick] (Fact1.east) -- (Revise1.west);
    \draw[-, thick] (Fact2.east) -- (Revise2.west);
    \draw[-, thick] (Fact3.east) -- (Revise3.west);
    \draw[-, thick] (Fact4.east) -- (Revise4.west);
    \draw[-, thick] (Fact5.east) -- (Revise5.west);
    
    \draw[-, thick] (Revise1.east) -- (Relevance1.west);
    \draw[-, thick] (Revise2.east) -- (Relevance2.west);
    \draw[-, thick] (Revise3.east) -- (Relevance3.west);
    \draw[-, thick] (Revise4.east) -- (Relevance4.west);
     \draw[-, thick] (Revise5.east) -- (Relevance5.west);
    
    \draw[-, thick] (Relevance1.east) -- (Rate1.west);
    \draw[-, thick] (Relevance2.east) -- (Rate2.west);
    \draw[-, thick] (Relevance3.east) -- (Rate3.west);
    \draw[-, thick] (Relevance4.east) -- (Rate4.west);
    \draw[-, thick] (Relevance5.east) -- (Rate5.west);

    \draw[-, thick] (Rate1.east) -- (Output.west);
    \draw[-, thick] (Rate2.east) -- (Output.west);
    \draw[-, thick] (Rate3.east) -- (Output.west);
    \draw[-, thick] (Rate4.east) -- (Output.west);
    \draw[-, thick] (Rate5.east) -- (Output.west);

\end{tikzpicture}
}
\caption{The RAFS assesses the factual accuracy of reasoning paths in four steps. (1) Split into sentences: The reasoning is divided into individual statements. (2) Generate retrieval queries: For each statement, an LLM generates multiple queries aimed at retrieving relevant information. (3) Retrieve information: The retrieval system gathers supporting information based on these queries. (4) Rate using retrieved information: Each statement is evaluated against the retrieved information and labeled as \textcolor{green}{Supported} or \textcolor{red}{Not Supported}. The final output includes a factuality score, calculated as the proportion of supported statements, which aids in selecting the most factually reliable reasoning path.}
    \label{fig:rafs}
\end{figure*}

\subsection{Retrieval-Augmented Generator}
Traditional retrieval-augmented generation methods often rely on a single retrieval step before generating responses, which limits their effectiveness in complex reasoning tasks. However, recent advances in medical RAG have demonstrated the importance of \textbf{iterative retrieval} in improving reasoning quality. i-MedRAG \cite{xiong2024improving} highlights that multi-turn retrieval, where an LLM generates follow-up queries dynamically, significantly enhances response accuracy by refining its knowledge base. Inspired by this approach, we introduce two new retrieval-augmented actions into the original rStar self-generator \cite{qi2024mutual}, transforming it into a \textbf{Retrieval-Augmented Generator}, as illustrated in Figures \ref{fig:a6} and \ref{fig:a7}. These retrieval-augmented actions enable the generator to dynamically incorporate external knowledge during intermediate reasoning steps, improving both the contextual relevance and factual accuracy of generated responses.

\begin{itemize}
    \item\textbf{A6: Search Query Generation and Information Retrieval}. In this step, the LLM generates targeted search queries from the original question and retrieves relevant information. This additional context is then used to enrich the reasoning path, enabling the model to deliver a more comprehensive and contextually grounded final answer. 
    \item\textbf{A7: Sub-question Retrieval and Re-answering}. Unlike Action A6, which centers on the main question, this action targets sub-questions generated by Action A3. For each sub-question, the model fetches relevant information and re-answers accordingly. By refining these intermediary steps, A7 improves the reasoning chain’s coherence and factual reliability, resulting in more accurate overall outcomes.
\end{itemize}

By incorporating these retrieval-augmented actions, the generator can explore a wider range of possible solutions, leading to reasoning paths that are both logically coherent and enriched with external knowledge.
This upgrade transforms the generator into a retrieval-augmented generator, enabling RARE to better handle complex, knowledge-intensive reasoning tasks. We draw on a diverse corpus—PubMed, StatPearls, medical textbooks, and Wikipedia—for authoritative, up-to-date information. For efficient retrieval, we employ the ColBERT model \cite{khattab2020colbert}, a dense retrieval approach optimized for passage-level retrieval, enabling fine-grained token-level matching to enhance the precision of retrieved information. Additional details on the retrieval corpus and model can be found in Appendix  ~\ref{subsec:a4}, the prompt for each action can be found in Appendix ~\ref{subsec:a7}.

\subsection{Retrieval-Augmented Factuality Scorer (RAFS)}

Inspired by the Search Augmented Factuality Evaluator (SAFE) \cite{wei2024long}, which combines an LLM (GPT-3.5-turbo) with Google Search to evaluate the factuality of responses, RARE introduces the \textbf{Retrieval-Augmented Factuality Scorer}. RAFS adapts this approach by replacing GPT-3.5-turbo with LLaMA 3.1 and Google Search with a corpus index retrieval system containing both general-domain knowledge (Wikipedia) and medical-domain resources (PubMed, StatPearls, and Medical Textbooks). This adaptation enhances domain specificity and accessibility for tasks requiring specialized knowledge. To assess the factual accuracy of generated reasoning paths, RAFS evaluates each candidate trajectory in four systematic steps, as illustrated in Figure \ref{fig:rafs}. More details about the retrieval corpus can be found in Appendix ~\ref{subsec:a4}.

\begin{enumerate}
    \item \textbf{Split into Statements}: Each reasoning path is divided into individual statements. 
    \item \textbf{Generate Retrieval Queries}: For each statement, RAFS employs an LLM to generate multiple retrieval queries designed to retrieve contextually relevant evidence. 
    \item \textbf{Retrieve Information}: The retrieval system gathers documents or information that corresponds to each generated query. 
     \item \textbf{Rate Using Retrieved Information}: Each statement is compared against the retrieved evidence and labeled as either \textbf{Supported} or \textbf{Not Supported}, based on alignment with the information. The overall factuality score for the reasoning path is calculated as the proportion of supported statements, indicating the trajectory’s factual reliability.
\end{enumerate}

As shown in Figure \ref{fig:rafs}, RAFS outputs a factuality score along with \textbf{Supported} or \textbf{Not Supported} labels for each statement. This scoring aids in selecting the most reliable reasoning path from multiple candidates, allowing RARE to prioritize responses that align closely with verified external knowledge.

\begin{table}
\resizebox{\linewidth}{!}{
\centering
\begin{tabular}{l|l|c|c|c|c}
\textbf{Model} & \textbf{Method} & \textbf{MedQA} & \textbf{MedMCQA} & \textbf{MMLU-M} & \textbf{Avg} \\ \hline
\multirow{6}{*}{LLaMA3.2 3B} 
  & CoT     & 52.63  & 49.82  & 57.67  & 53.37 \\
  & MedRAG  & 52.08  & 51.78  & 65.58  & 56.48 \\
  & i-MedRAG& 60.88  & 53.60  & 66.76  & 60.41 \\
  & SC      & 56.09  & 50.85  & 58.49  & 55.14 \\
  & rStar   & 61.27  & 54.26  & 67.22  & 60.92 \\
  & RARE    & \textbf{63.86} & \textbf{56.61} & \textbf{70.98} & \textbf{63.82} \\ \hline
\multirow{6}{*}{LLaMA3.1 8B} 
  & CoT     & 61.51  & 55.15  & 71.63  & 62.76 \\
  & MedRAG  & 63.00  & 56.87  & 74.56  & 64.81 \\
  & i-MedRAG& 73.61  & 61.61  & 78.42  & 71.21 \\
  & SC      & 64.73  & 56.35  & 72.73  & 64.60 \\
  & rStar   & 70.40  & 62.13  & 79.16  & 70.56 \\
  & RARE    & \textbf{75.57} & \textbf{64.32} & \textbf{81.63} & \textbf{73.84} \\ \hline
\multirow{6}{*}{LLaMA3.1 70B} 
  & CoT     & 76.67  & 68.75  & 81.72  & 75.71 \\
  & MedRAG  & 77.61  & 71.19  & 84.76  & 77.85 \\
  & i-MedRAG& 82.40  & 72.38  & 86.69  & 80.49 \\
  & SC      & 79.49  & 70.19  & 82.73  & 77.47 \\
  & rStar   & 84.99  & 72.72  & 87.19  & 81.63 \\
  & RARE    & \textbf{87.43} & \textbf{75.18} & \textbf{90.91} & \textbf{84.51} \\ \hline
Meditron 70B & CoT & 51.69 & 46.74 & 64.92 & 54.45 \\ \hline
Mixtral (8x7B) & CoT & 64.10 & 56.28 & 74.01 & 64.80 \\ \hline
GPT-3.5 & CoT & 65.04 & 55.25 & 72.91 & 64.40 \\ \hline
GPT-4 & CoT & 83.97 & 69.88 & 89.44 & 81.10 \\ \hline
GPT-4o Mini & CoT & 73.29  & 66.17  & 84.30  & 74.59 \\ \hline
GPT-4o & CoT & 85.55 & 74.70 & 90.45 & 83.57 \\ 
\end{tabular}
}
\caption{Performance of RARE and baseline methods on three medical reasoning benchmarks: MedQA, MedMCQA, and MMLU-Medical. SC is self-consistency. RARE consistently outperforms rStar across all model sizes, with improvements statistically significant at \textbf{p < 0.01} based on paired t-tests over multiple runs.
}
\label{tab:medical_performance}
\end{table}

\section{Results}
In this section, we evaluate the performance of our proposed method, RARE, on both medical reasoning and commonsense reasoning tasks using three large language models: LLaMA 3.2 3B Instruct, LLaMA 3.1 8B Instruct and LLaMA 3.1 70B Instruct \cite{dubey2024LLaMA}. Throughout our work, we may drop ``Instruct'', but we are always referring to the ``Instruct'' versions. Detail settings of the experiments, descriptions of the evaluation tasks and baselines can be found in Appendix.

\subsection{Performance on Medical Reasoning tasks}

Table \ref{tab:medical_performance} shows the performance of RARE and various baseline methods on three challenging medical reasoning benchmarks: MedQA, MedMCQA, and MMLU-Medical. These datasets require not only complex reasoning but also a high degree of factual accuracy, making them suitable for evaluating the effectiveness of RARE’s retrieval-augmented reasoning approach. 
The results demonstrate the effectiveness of RARE in enhancing the reasoning capabilities of LLaMA models compared to baseline methods, including Chain of Thought,  MedRAG, i-MedRAG, Self-Consistency(SC), and rStar. 
Across all model sizes—LLaMA3.2 3B, LLaMA3.1 8B, and LLaMA3.1 70B—RARE consistently outperforms baseline methods. 
On LLaMA3.1 8B, RARE achieves substantial gains, outperforming rStar by 5.17\% on MedQA, 2.19\% on MedMCQA, and 2.47\% on MMLU-Medical. The performance improvement becomes more pronounced as model size increases, with RARE-enhanced LLaMA3.1 70B outperforming GPT-4o on MedQA (87.43\% vs. 85.55\%) and MMLU-Medical (90.91\% vs. 90.45\%), highlighting its competitive edge. 


\subsection{Performance on Commonsense Reasoning}

Table \ref{tab:commonsense_comparison} presents the performance of RARE compared to other methods and larger language models on commonsense reasoning benchmarks, including StrategyQA, CommonsenseQA, and Physical IQA. These datasets test a range of commonsense reasoning skills, with StrategyQA requiring more multi-step reasoning. RARE consistently outperforms baseline methods, including CoT, MedRAG, iMedRAG, SC and rStar, across both LLaMA3.1 8B and LLaMA3.1 70B models. For LLaMA3.1 8B, RARE achieves substantial improvements over rStar, with gains of 6.45\% on StrategyQA, 4.26\% on CommonsenseQA, and 2.87\% on PIQA. On LLaMA3.1 70B, RARE further closes the gap with state-of-the-art proprietary models, achieving 85.74\% on StrategyQA, 86.98\% on CommonsenseQA, and 92.66\% on PIQA, surpassing GPT-4o.


RARE’s retrieval-augmented reasoning method provides substantially larger gains over CoT on multi-step inference tasks (like StrategyQA) compared to commonsense-heavy tasks (such as CommonsenseQA), indicating its particular effectiveness in handling implicit, multi-hop reasoning. Specifically, RARE’s largest improvement over CoT is observed on StrategyQA(10.19\%), where multi-step reasoning is crucial, suggesting that its retrieval-augmented reasoning enhancement method is particularly effective in handling implicit, multi-hop inference. In contrast, while RARE outperforms CoT on CommonsenseQA(7.22\%), the gains are relatively smaller, indicating that tasks relying more on commonsense knowledge rather than explicit step-by-step reasoning do not benefit as significantly from retrieval-augmented reasoning.

\subsection{Ablation Study}

\begin{table}
\centering
\resizebox{\linewidth}{!}{
\begin{tabular}{l|c|c|c|c|c}
\textbf{Model} & \textbf{Method} & \textbf{SQA} & \textbf{CQA} & \textbf{PIQA} & \textbf{Avg} \\ \hline
\multirow{5}{*}{LLaMA3.1 8B} 
  & CoT     & 67.83  & 73.62  & 76.17  & 72.54 \\
  & MedRAG  & 66.08  & 74.45  & 78.67  & 73.07 \\
  & i-MedRAG& 68.12  & 75.18  & 80.52  & 74.61 \\
  & SC      & 68.41  & 74.90  & 77.42  & 73.58 \\
  & rStar   & 71.57  & 76.58  & 83.04  & 77.06 \\
  & RARE    & \textbf{78.02} & \textbf{80.84} & \textbf{85.91} & \textbf{81.59} \\ \hline
\multirow{5}{*}{LLaMA3.1 70B} 
  & CoT     & 76.71  & 78.62  & 81.66  & 79.00 \\
  & MedRAG  & 75.54  & 82.23  & 86.07  & 81.28 \\
  & i-MedRAG& 77.29  & 83.13  & 87.76  & 82.73 \\
  & SC      & 77.29  & 78.87  & 82.67  & 79.61 \\
  & rStar   & 81.80  & 83.66  & 89.27  & 84.91 \\
  & RARE    & \textbf{85.74} & \textbf{86.98} & \textbf{92.66} & \textbf{88.46} \\ \hline
Claude-3 Haiku & CoT & 69.58 & 67.40 & 82.32 & 73.10 \\ \hline
Claude-3.5 Sonnet & CoT & 76.86 & 74.12 & 89.39 & 80.12 \\ \hline
GPT-4o Mini & CoT & 78.60 & 82.31 & 88.41 & 83.11 \\ \hline
GPT-4o & CoT & 80.64 & 86.50 & 91.13 & 86.09 \\ 
\end{tabular}
}
\caption{Performance comparison on common sense reasoning tasks with various LLMs and reasoning methods, evaluated on StrategyQA (SQA), CommonsenseQA (CQA), and Physical IQA (PIQA). SC is self-consistency. RARE consistently outperforms rStar across all model sizes, with improvements statistically significant at \textbf{p < 0.01} based on paired t-tests over multiple runs.}
\label{tab:commonsense_comparison}
\end{table}

\begin{table}[h!]
\centering
\resizebox{\linewidth}{!}{
\begin{tabular}{l|c}
\textbf{Configuration} & \textbf{Accuracy} \\ \hline
rStar & 70.0 \\ 
rStar + RAFS & 71.4 \\ 
rStar + A6 & 72.4 \\ 
rStar + A7 & 71.2 \\ 
rStar + A6 + A7 & 73.2 \\ 
RARE (rStar + A6 + A7 + RAFS) & \textbf{74.8} \\ 
\end{tabular}}
\caption{Ablation study on RARE components, evaluated on 250 MedQA samples using LLaMA 3.1 8B.}
\label{tab:ablation_study}
\end{table}

We conduct an ablation study on 250 MedQA samples using the LLaMA 3.1 8B model to assess the contributions of each RARE component (Table~\ref{tab:ablation_study}). Starting with the baseline (rStar) at 70.0\% accuracy, adding the RAFS improves reliability, increasing accuracy to 71.4\%. Incorporating A6 and A7 further boosts accuracy to 72.4\% and 71.2\%, respectively. Combining both actions yields 73.2\%, demonstrating their synergy in strengthening reasoning. The full RARE configuration, integrating rStar, A6, A7, and RAFS, achieves the highest accuracy (74.8\%), highlighting the collective impact of retrieval and factuality scoring in enhancing reasoning accuracy.

\subsection{Common Reasoning Paths Patterns}
Exploring common paths leading to correct answers helps us understand and refine the reasoning patterns that most reliably produce accurate solutions. Figures \ref{fig:medbar} and \ref{fig:stgbar} illustrate the top 10 most common reasoning paths that lead to correct answers on MedQA and StrategyQA, respectively.  

In MedQA, prominent paths such as \texttt{A1 $\rightarrow$ A2}, \texttt{A3 $\rightarrow$ A2}, and \texttt{A1 $\rightarrow$ A6} constitute a large share of successful reasoning. More complex sequences like \texttt{A3 $\rightarrow$ A7 $\rightarrow$ A3} also appear, demonstrating the generator’s capacity to explore multiple approaches adaptively.

A similar pattern occurs on StrategyQA, where paths like \texttt{A1 $\rightarrow$ A2}, \texttt{A3 $\rightarrow$ A2}, and \texttt{A1 $\rightarrow$ A6} remain dominant. Still, simpler actions such as \texttt{A6} and \texttt{A2} also play a noticeable role, reflecting the distinct reasoning demands of this task.

Overall, both figures highlight the RARE's flexibility in navigating a variety of reasoning strategies. They further illustrate the importance of both straightforward and more intricate paths in effectively tackling different categories of tasks.

\subsection{Human Evaluation of the Retrieval-Augmented Factuality Scorer}

\begin{table}[h]
\centering
\resizebox{\linewidth}{!}{
\begin{tabular}{l|c}
\textbf{Metric} & \textbf{Score (\%)} \\ \hline
Inter-Annotator Agreement & 86.49 \\
RAFS Alignment with Annotator 1 & 87.84 \\
RAFS Alignment with Annotator 2 & 82.43 \\
Average RAFS-Annotator Alignment & \textbf{85.14} \\ 
\end{tabular}
}
\caption{Human evaluation of RAFS.}
\label{tab:rafs_human_eval}
\end{table}

To assess the effectiveness of RAFS, we conducted a human evaluation comparing its factual assessments with expert judgments. We selected 10 medical questions from MedQA and generated corresponding responses using RARE with LLaMA 3 8B, segmenting each response into approximately 100 individual statements. RAFS classified each statement as either \textbf{Supported} or \textbf{Not Supported}. Two medical experts—both of whom have passed the United States Medical Licensing Examination (USMLE)—independently labeled each statement as \textbf{Correct} or \textbf{Incorrect}, corresponding to RAFS’s \textbf{Supported} and \textbf{Not Supported} labels, respectively.

Table~\ref{tab:rafs_human_eval} shows a high inter-annotator agreement of 86.49\%, confirming consistency among expert evaluations. RAFS achieves strong alignment with human judgments, with an average agreement of 85.14\% (87.84\% with Annotator 1 and 82.43\% with Annotator 2), demonstrating its reliability in factual assessment. The slightly lower alignment with Annotator 2 suggests some subjectivity in evaluating borderline statements. Overall, RAFS effectively prioritizes factually accurate reasoning paths, closely aligning with expert validation.


\subsection{Computational Analysis}
\begin{figure}[!ht]
    \centering
    \includegraphics[width=1.0\linewidth]{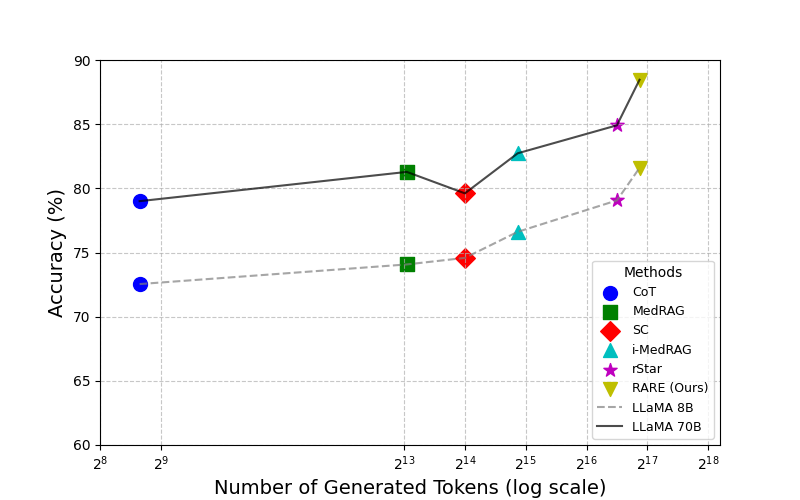}
    \caption{Accuracy vs. Number of Generated Tokens for Different Methods}
    \label{fig:comp}
\end{figure}
Figure~\ref{fig:comp} illustrates the trade-off between inference cost and reasoning performance, averaged across commonsense reasoning tasks. CoT requires approximately 400 tokens per example, while rStar averages 92.6k tokens, and RARE increases this to 119.9k tokens due to its additional retrieval-augmented steps. On average, rStar improves accuracy over CoT by 4.52\%, and RARE achieves a 9.05\% improvement—the highest among all methods. Although RARE uses roughly 27.3k more tokens than rStar, its superior accuracy demonstrates that the additional computation enhances reasoning robustness. This finding aligns with recent trends such as OpenAI O1 and DeepSeek R1 \cite{el2025competitive, guo2025deepseek}, which prioritize deeper reasoning over minimal latency, reinforcing that improved accuracy can justify increased computational cost in complex reasoning tasks.

\section{Related Work}

\subsection{Prompting LLMs to reason} 
Prompting LLMs to reason has become a central focus of recent research, particularly through the development of prompting-based techniques such as Chain-of-Thought prompting~\cite{wei2022chain}.
These approaches aim to improve inference capabilities by structuring prompts to encourage multi-step reasoning. Key advancements include planning-based prompting~\cite{hao2023reasoning, ding2023everything}, problem decomposition~\cite{zhou2022least, khot2022decomposed}, abstraction~\cite{zheng2023take}, and program-based reasoning~\cite{chen2022program, zhou2023solving}. While effective for single-step inference, these methods often rely on a limited range of operations, which can restrict the diversity and generality of the solutions generated.

MCTS has emerged as a powerful tool for optimizing reasoning paths across a vast solution space, enhancing both exploration and decision-making efficiency~\cite{silver2018general}. MCTS has been widely adopted in domains such as game theory~\cite{sironi2018self, ontanon2016informed} and strategic planning~\cite{zhou2023language, yu2023prompt}. When integrated with reinforcement learning, MCTS enables self-play training, achieving human-level or even superhuman performance in complex domains like Go~\cite{silver2016mastering}.

More recently, MCTS has been applied to LLMs to identify optimal reasoning trajectories without requiring additional supervision~\cite{feng2023alphazero, zhou2023language, tian2024toward, huang2024ritek}. For instance, \citet{feng2023alphazero} restricted the search granularity to word or sentence level, while \citet{tian2024toward} introduced $\eta$MCTS to enable hierarchical planning with tailored reward functions. MCTS has also been employed to curate high-quality reasoning paths for training reward models, leading to iterative improvements in model reasoning~\cite{zhang2024rest}. \citet{zhou2023language} further generalized the use of MCTS by incorporating all possible reasoning and action steps into a unified search space, supporting joint inference, action, and planning.

The most relevant work to ours is rStar~\cite{qi2024mutual}, which extends MCTS by incorporating a diverse set of reasoning operations. rStar improves solution-space exploration through dynamic, human-like reasoning and introduces five distinct operations that enable more flexible planning. Moreover, it employs a discriminator—another LLM with reasoning capabilities similar to the main model—to verify the plausibility of each candidate reasoning path.

\subsection{Medical RAG} 

Retrieval-augmented generation has proven highly effective in grounding LLM reasoning with up-to-date external knowledge, especially in medical tasks such as question answering and generation~\cite{xiong2024benchmarking, tian2019chimed, xia2022medconqa, wang2024jmlr, yao2024mcqg, yao2025survey}. RAG has also been applied to classification, information extraction, lay-language generation~\cite{li2024biomedrag, guo2024retrieval, yao2023readme}, and medical dialogue systems~\cite{shi2024medical}.
To enhance retrieval effectiveness, various improvements have been proposed, such as query rewriting~\cite{zhang2022focus} and multi-step retrieval~\cite{mrini2022medical}, enabling LLMs to iteratively refine their knowledge intake for clinical decision-making and literature synthesis~\cite{zakka2024almanac, wang2023augmenting, jeong2024improving}. For example, \citet{wang2023augmenting} developed a hybrid retriever with complex filtering mechanisms to identify high-quality documents, while \citet{jeong2024improving} introduced SelfBioRAG, a self-reflective retriever that integrates reasoning signals into the retrieval loop.
Iterative RAG variants like i-MedRAG allow LLMs to refine their understanding by posing follow-up queries in multiple rounds~\cite{xiong2024improving}. The most relevant RAG-based system to our work is SeRTS~\cite{hu2024serts}, which leverages MCTS to guide query generation. However, SeRTS follows a depth-first strategy, generating and executing one query at a time, whereas our framework—RARE—adopts a more flexible, agentic planning approach.

RARE integrates both RAG-based and non-RAG operations, allowing for breadth-first exploration via A6, which generates multiple queries simultaneously. This enables RARE to handle a wider variety of reasoning tasks. In contrast to SeRTS, which always requires iterative retrieval, RARE can also answer simple problems directly via step-by-step (CoT-like) reasoning without invoking retrieval. This design makes RARE more versatile, allowing it to dynamically adapt between retrieval-driven and retrieval-free reasoning depending on the task complexity.

\section{Conclusion}

We introduced RARE (Retrieval-Augmented Reasoning Enhancement), a framework designed to improve the reasoning accuracy and factual reliability of large language models (LLMs) through retrieval-augmented actions and factuality scoring. RARE operates entirely as an autonomous language agent, requiring no additional training or fine-tuning of the underlying LLM. This makes the framework robust to overfitting and highly adaptable across tasks and datasets, as it relies solely on real-time retrieval and reasoning mechanisms. Experiments on medical and commonsense reasoning benchmarks demonstrate RARE’s effectiveness, RARE bridges the gap between open-source models and state-of-the-art proprietary systems, showcasing its potential as a scalable and effective solution for knowledge-intensive reasoning tasks.


\section{Limitations}

While RARE demonstrates significant improvements in reasoning accuracy and factual reliability, it has certain limitations that present opportunities for future work.

First, the framework has only been tested on open-source models like LLaMA 3.1 and not on larger proprietary models such as GPT-4. This is due to the high number of API calls required by RARE's iterative retrieval and reasoning process, making evaluations on closed-source models prohibitively costly. However, the framework is designed to be model-agnostic and can be directly applied to proprietary models if resources permit.

Second, RARE is designed to identify a single reasoning path that leads to a correct answer but does not necessarily optimize for the best or shortest path that maximizes robustness (e.g., achieving the highest model confidence). Future work could explore designing better reward functions to prevent reward hacking and improve the selection of the most reliable reasoning paths.


Finally, RARE is currently limited to using Monte Carlo Tree Search for exploring action paths. While effective, this approach does not leverage a trained reward model to dynamically guide the search process. Future extensions could incorporate reward models or alternative optimization strategies to further enhance reasoning quality and efficiency.

These limitations highlight areas for improvement and potential research directions to make RARE more robust, generalizable, and applicable to a wider range of models and reasoning tasks.

\section{Ethics Statement}

This work aims to advance the field of Medical QA by enhancing the reasoning capabilities of language models through the RARE framework. While the results demonstrate significant improvements, several ethical considerations must be addressed to ensure responsible development and deployment:

\textbf{Clinical Applicability.} RARE is designed to improve reasoning and factual reliability but is not intended to replace healthcare professionals or serve as a standalone diagnostic or treatment tool. Any integration into medical workflows must be supervised by qualified practitioners to ensure patient safety and ethical use.

\textbf{Bias and Fairness.} Language models, including those tested with RARE, may reflect biases present in their training data. These biases could impact the fairness and reliability of the reasoning process, particularly in sensitive medical contexts. Future work must include rigorous audits for bias and fairness to minimize potential harm.

\textbf{Generalizability.} As RARE has been primarily evaluated in English-language, text-based general and medical domain QA tasks, its applicability to non-English-speaking contexts and multimodal scenarios remains untested. Efforts should be made to extend the framework to diverse linguistic and cultural contexts to ensure equitable access to its benefits.

\textbf{Societal Impacts.} While RARE demonstrates the potential for improving medical reasoning tasks, its outputs should be considered supplementary to human expertise. The ethical deployment of RARE requires clear guidelines to avoid overreliance on AI and ensure that it enhances, rather than replaces, human decision-making in healthcare.

\textbf{Human Annotation.} The human evaluation in this study was conducted by two medical experts who have passed the United States Medical Licensing Examination (USMLE). Annotators were compensated fairly for their time at a rate of \$40/hour. All participants were informed about the scope of their role and participated voluntarily.

\bibliography{acl25}
\bibliographystyle{acl_natbib}

\newpage

\appendix
\include{Appendix}
\section{Appendix}
\label{sec:appendix}

\subsection{Implementation Details}

In the MCTS framework of the Retrieval-augmented Generator, we set the number of rollouts to 4 for LLaMA 3.2 3B and LLaMA 3.1 8B models, and 2 for the LLaMA 3.1 70B model. This configuration strikes a balance between effective inference and computational efficiency, particularly for larger models where inference costs are higher.

In the factuality scoring stage, we perform a self-scorer setup, where the Retrieval-Augmented Factuality Scorer uses the same backbone model as the generator. For instance, when the generator utilizes LLaMA 3.1 3B, the RAFS also employs the LLaMA 3.2 3B model for factuality evaluation. This ensures consistency between the generator and scorer while maintaining efficient inference. All inference processes, including factuality scoring, are parallelized to further enhance efficiency.





\subsection{Evaluation tasks}

To rigorously test the reasoning capabilities of RARE, we evaluate it on a range of reasoning tasks, categorized into two main domains:

\textbf{Medical Reasoning Tasks}: We use three medical datasets that require complex, domain-specific reasoning, including:
    \begin{itemize}
        \item \textbf{MedQA-USMLE} \cite{jin2021disease}: A medical question-answering dataset based on the United States Medical Licensing Examination (USMLE) questions.
        \item \textbf{MedMCQA} \cite{pal2022medmcqa}: A dataset consisting of multiple-choice medical questions designed to test clinical knowledge.
        \item \textbf{MMLU-Medical} \cite{singhal2023large}: The medical subset of the Massive Multitask Language Understanding (MMLU) benchmark, focusing on diverse topics in the medical field.
    \end{itemize}

\textbf{Commonsense Reasoning Tasks}: We evaluate RARE’s general reasoning ability on commonsense datasets. While \textbf{StrategyQA} requires complex, implicit reasoning strategies, other tasks benefit from advanced reasoning but may not require it to the same extent:
    \begin{itemize}
        \item \textbf{StrategyQA} \cite{geva2021did}: A dataset of open-domain questions that require implicit reasoning strategies.
        \item \textbf{CommonsenseQA} \cite{talmor2018commonsenseqa}: A multiple-choice question-answering dataset designed to test commonsense knowledge.
        \item \textbf{PIQA (Physical Interaction QA)} \cite{bisk2020piqa}: A dataset for physical reasoning, where models must answer questions about common physical interactions.
    \end{itemize}

\subsection{Baselines}

We compare the performance of RARE with several baseline reasoning methods, including:
\begin{itemize}
    \item \textbf{Chain of Thought (CoT)} \cite{wei2022chain}: A reasoning approach that generates explanations step-by-step, aiming for more coherent answers.
    \item \textbf{Self-Consistency} \cite{wang2022self}: A method that uses majority voting among multiple reasoning paths to increase response accuracy.
    \item \textbf{rStar} \cite{qi2024mutual}: A framework that extends MCTS with a diverse set of reasoning actions, improving reasoning accuracy.
    \item \textbf{MedRAG} \cite{xiong2024benchmarking}: A retrieval-augmented framework designed for medical question answering, integrating knowledge retrieval from domain-specific corpora.
    \item \textbf{i-MedRAG} \cite{xiong2024improving}: An iterative retrieval-based medical reasoning model that dynamically refines queries to improve response accuracy in medical QA tasks.
    
\end{itemize}




\subsection{Retrieval Model and Corpus}
\label{subsec:a4}

\begin{table}[ht]
\centering
\scalebox{0.85}{
\begin{tabular}{l|r|r|r|l}
\toprule
\textbf{Corpus} & \textbf{\#Docs} & \textbf{\#Snippets} & \textbf{Avg. L} & \textbf{Domain} \\ 
\midrule
PubMed       & 23.9M  & 23.9M  & 296  & Biomed.   \\ 
StatPearls   & 9.3k   & 301.2k & 119  & Clinics   \\ 
Textbooks    & 18     & 125.8k & 182  & Medicine  \\ 
Wikipedia    & 6.5M   & 29.9M  & 162  & General   \\ 
MedCorp      & 30.4M  & 54.2M  & 221  & Mixed     \\ 
\bottomrule
\end{tabular}
}
\caption{Statistics of the retrieval corpus used in our experiments. \#Docs refers to the number of documents, \#Snippets represents extracted text units, and Avg. L indicates the average snippet length.}
\label{tab:retrieval_corpus}
\end{table}

For information retrieval, we leverage the \textbf{MedCorp} \cite{xiong2024benchmarking} corpus, a curated collection of high-quality and domain-specific resources that serve as knowledge bases for medical and general question answering. The \textbf{MedCorp} corpus consists of (table \ref{tab:retrieval_corpus}):

\begin{itemize}
    \item \textbf{PubMed}\footnote{https://pubmed.ncbi.nlm.nih.gov/}: A widely used biomedical literature database containing over 36 million articles~\cite{lu2011pubmed, jin2024pubmed}. For our retrieval tasks, we utilize a subset of \textbf{23.9 million} articles with valid titles and abstracts, similar to the MedRAG setup~\cite{xiong2024benchmarking}.
    \item \textbf{StatPearls}\footnote{https://www.statpearls.com/}: A clinical decision support resource with publicly available medical articles hosted on \textbf{NCBI Bookshelf}\footnote{https://www.ncbi.nlm.nih.gov/books/NBK430685/}. The corpus includes \textbf{9,330} peer-reviewed StatPearls articles, structured into hierarchical snippets where each paragraph is treated as a retrieval unit, with corresponding hierarchical headings as metadata.
    \item \textbf{Medical Textbooks}\footnote{https://github.com/jind11/MedQA}: A collection of \textbf{18 widely used medical textbooks}~\cite{jin2021disease} that are commonly referenced for foundational medical knowledge and United States Medical Licensing Examination (USMLE) preparation. The textbook corpus is segmented into passages of up to \textbf{1,000 characters} using the RecursiveCharacterTextSplitter from LangChain\footnote{https://www.langchain.com/}.
    \item \textbf{Wikipedia}: A large-scale, general-domain encyclopedia frequently used in information retrieval tasks~\cite{thakur2021beir}. We incorporate a \textbf{processed version} of Wikipedia from Huggingface\footnote{https://huggingface.co/datasets/wikipedia} and apply text chunking techniques to facilitate passage-level retrieval.
\end{itemize}

To retrieve relevant passages from \textbf{MedCorp}, we employ the \textbf{ColBERT retrieval model}~\cite{khattab2020colbert}, an efficient neural retrieval framework optimized for \textbf{passage ranking}. ColBERT’s \textbf{late interaction mechanism} enables fine-grained token-level relevance matching, allowing our retrieval-augmented generator to identify and incorporate \textbf{contextually relevant} medical knowledge. This approach enhances the factual accuracy and reliability of generated responses while efficiently handling large-scale biomedical corpora.

\subsection{Monte Carlo Tree Search (MCTS)}
\label{subsec:a5}

Monte Carlo Tree Search (MCTS) is a decision-making algorithm widely used in complex decision processes, such as games, by building a search tree and simulating outcomes to estimate the value of potential actions. MCTS operates through four main phases \cite{browne2012survey}. \\ 
     \textbf{Selection}: Starting from the root node, the algorithm traverses through child nodes based on strategies like the Upper Confidence Bound applied on Trees (UCT), which balances exploration and exploitation, continuing until a leaf node is reached. \\
     \textbf{Expansion}: At the leaf node, if it does not represent a terminal state, one or more feasible child nodes are added to represent potential future actions. \\
     \textbf{Simulation (Evaluation)}: From one of the newly added nodes (typically selected randomly), random simulations (or "rollouts") are performed by selecting actions randomly until reaching a terminal state, thereby estimating the node's potential. \\
     \textbf{Backpropagation}: After simulation, the results (win, loss, or draw) are propagated back through the traversed nodes, updating the statistical data (e.g., rewards or visit counts) to guide future decision-making.

By iterating through these phases, MCTS incrementally builds a decision tree, enabling optimal strategy refinement in scenarios where direct calculation of the best strategy is infeasible due to a vast state space. A crucial component of MCTS is the Upper Confidence Bound applied on Trees (UCT) algorithm, used during the selection phase to balance exploration and exploitation. The UCT formula for choosing actions is defined as follows:
\vspace{-2mm}
\begin{equation}
\text{UCT}_j = \bar{X}_j + C \sqrt{\frac{2 \ln N}{N_j}}
\end{equation}

where $\bar{X}_j$ is the average reward of action $j$, $N$ is the total visit count of the parent node, $N_j$ is the visit count of node $j$, and $C$ is a constant that controls the balance between exploration and exploitation.

rStar enhanced MCTS-based exploration of candidate solutions. Specifically, rStar leverages a reward mechanism to guide tree expansion. Each node $s$ generated under action $a$ has a reward value $Q(s, a)$. Unexplored nodes are initialized with $Q(s_i, a_i) = 0$, leading to random tree expansions initially. Upon reaching a terminal node $s_d$, a reward score $Q(s_d, a_d)$ is computed based on whether the trajectory reaches the correct answer and is then back-propagated to each intermediate node in the trajectory $t = x \oplus s_1 \oplus s_2 \oplus ... \oplus s_d$. For each intermediate node $s_i$, the reward is updated as $Q(s_i, a_i) = Q(s_i, a_i) + Q(s_d, a_d)$, with $Q(s_d, a_d)$ determined by the likelihood or confidence from self-consistency majority voting. This reward propagation directs the search toward paths with higher probabilities of correctness, enhancing decision-making within the reasoning process.

\subsection{Evaluation of Reasoning Candidates with RAFS}

\begin{figure*}[ht]
\begin{tcolorbox}[colback=white, colframe=black, width=\textwidth, title=RAFS Examples]
\textbf{Question:} A 35-year-old man comes to the physician because of itchy, watery eyes for the past week. He has also been sneezing multiple times a day during this period. He had a similar episode 1 year ago around springtime. He has iron deficiency anemia and ankylosing spondylitis. Current medications include ferrous sulfate, artificial tear drops, and indomethacin. He works as an elementary school teacher. His vital signs are within normal limits. Visual acuity is 20/20 without correction. Physical examination shows bilateral conjunctival injection with watery discharge. The pupils are 3 mm, equal, and reactive to light. Examination of the anterior chamber of the eye is unremarkable. Which of the following is the most appropriate treatment? A: Erythromycin ointment, B: Ketotifen eye drops, C:Warm compresses, D: Fluorometholone eye drops

\vspace{0.2cm}
\textbf{Candidate Reasoning 1:} \textcolor{green}{Let's think step by step. Based on the symptoms described - itchy, watery eyes, sneezing, bilateral conjunctival injection, and watery discharge - along with the history of a similar episode around springtime, it's clear that the patient is experiencing allergic conjunctivitis, likely triggered by seasonal allergens like pollen given his profession as an elementary school teacher. For symptomatic relief of allergic conjunctivitis, antihistamine eyedrops or mast cell stabilizers can be effective. Among the options provided, ketotifen eye drops act as a mast cell stabilizer and would be appropriate for treating allergic conjunctivitis. The answer is B: Ketotifen eye drops.}  \(\checkmark\)  \textbf{The factuality score is 1.0}

\vspace{0.2cm}
\textbf{Candidate Reasoning 2:} \textcolor{green}{Based on the patient's symptoms of itchy, watery eyes, sneezing, and physical examination findings of bilateral conjunctival injection with watery discharge, the diagnosis is allergic conjunctivitis. Considering the timing of the episode and the patient's profession, the most likely causative agent is seasonal pollen. Treatment should focus on reducing symptoms rather than eliminating the allergen source. Fluorometholone eye drops are corticosteroid drops that reduce inflammation and can provide relief from itching and redness.} \textcolor{red}{ Other options like erythromycin or ketotifen may have some effect but would be less effective in addressing this presentation.} \textcolor{green}{Warm compresses might help with discharge, but again, would be less effective compared to the impact of corticosteroids.} \textcolor{red}{Therefore, the most appropriate treatment is fluorometholone eye drops. The answer is D: Fluorometholone eye drops.} \textcolor{red}{\(\times\)} \textbf{The factuality score is 0.625}

\vspace{0.2cm}
\textbf{Candiate Reasoning 3:} \textcolor{green}{Given the patient's symptoms of itchy, watery eyes, sneezing, and conjunctival injection, along with a history of similar episodes around springtime, this case is most consistent with seasonal allergic conjunctivitis. The best treatment for mild allergic conjunctivitis involves avoiding triggers when possible, using lubricating artificial tears regularly, and applying a cold compress. Warm compresses are often recommended to help loosen crusts and debris, improving comfort.} \textcolor{red}{Therefore, warm compresses would be the most appropriate treatment for this patient. The answer is C: Warm compresses.} \textcolor{red}{\(\times\)} \textbf{The factuality score is 0.6}

\end{tcolorbox}

\caption{
Evaluation of reasoning candidates by the Retrieval-Augmented Factuality Scorer (RAFS). Each candidate provides a reasoning path for a given question, with RAFS assigning a factuality score based on alignment with retrieved evidence.} 
\label{fig:rafs_example}

\end{figure*}

Figure \ref{fig:rafs_example} demonstrates how the Retrieval-Augmented Factuality Scorer (RAFS) evaluates reasoning candidates for a medical question. The example involves diagnosing and treating a patient presenting with symptoms consistent with seasonal allergic conjunctivitis. RAFS assigns factuality scores to each candidate reasoning path based on their alignment with retrieved evidence.

\textbf{Candidate Reasoning 1} accurately identifies the condition and treatment, correctly selecting ketotifen eye drops (Answer B) with a factuality score of 1.0.
\textbf{Candidate Reasoning 2} suggests fluorometholone eye drops (Answer D) but includes conflicting reasoning and inaccuracies, resulting in a lower factuality score of 0.625.
\textbf{Candidate Reasoning 3} incorrectly recommends warm compresses (Answer C) as the most appropriate treatment, reflecting limited factual alignment and receiving a score of 0.6.

This example highlights RAFS’s ability to differentiate between reasoning paths based on factual correctness, ensuring that the most reliable and evidence-supported answer is prioritized.

\begin{figure*}
    \centering
    \includegraphics[width=0.9\linewidth]{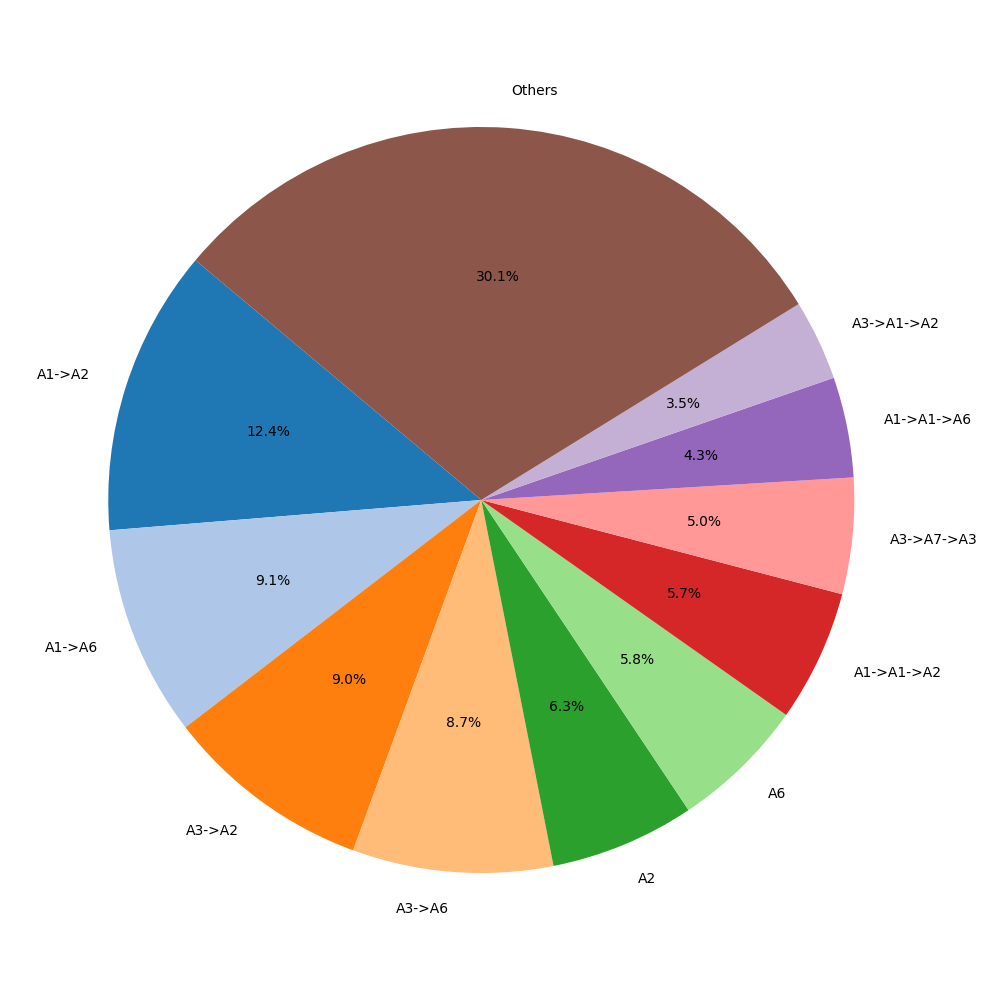}
    \caption{Top 10 common paths that lead to correct answer (MedQA)}
    \label{fig:medbar}
\end{figure*}

\begin{figure*}
    \centering
    \includegraphics[width=0.9\linewidth]{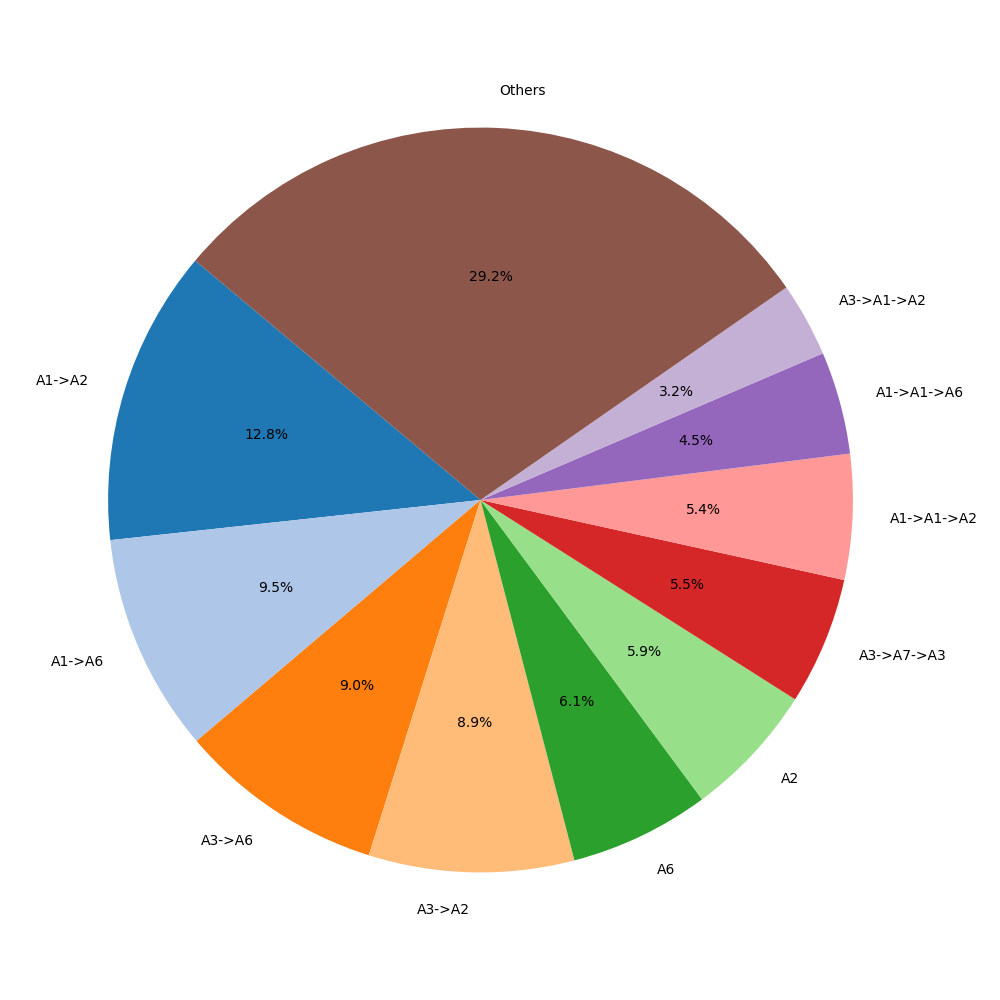}
    \caption{Top 10 common paths that lead to correct answer (StrategyQA)}
    \label{fig:stgbar}
\end{figure*}

\begin{figure*}[ht]
\begin{tcolorbox}[colback=white, colframe=black, width=\textwidth, title=Multi-step reasoning in Medical QA]
\textbf{Question:} A 23-year-old pregnant woman at 22 weeks gestation presents with burning upon urination. She states it started 1 day ago and has been worsening despite drinking more water and taking cranberry extract. She otherwise feels well and is followed by a doctor for her pregnancy. Her temperature is 97.7°F (36.5°C), blood pressure is 122/77 mmHg, pulse is 80/min, respirations are 19/min, and oxygen saturation is 98\% on room air. Physical exam is notable for an absence of costovertebral angle tenderness and a gravid uterus. Which of the following is the best treatment for this patient? A: Ampicillin, B: Ceftriaxone, C:Ciprofloxacin, D: Doxycycline, E: Nitrofurantoin

\vspace{0.1cm}
\textbf{Step 1: Identify key clinical features} A 23-year-old pregnant woman at 22 weeks gestation presents with burning upon urination (dysuria), worsening symptoms over one day, and no systemic signs of infection (normal temperature, blood pressure, and oxygen saturation). Physical exam shows no costoverterbral angle tenderness, suggesting the absence of pyelonephritis.

\vspace{0.1cm}
\textbf{Step 2: Determine the likely diagnosis} Dysuria without systemic signs in a pregnant woman is most consistent with acute cystitis, a lower urinary tract infection.

\vspace{0.1cm}

\textbf{Step 3: Evaluate treatment options for cystitis in pregnancy} Nitrofurantoin: Safe and effective during the second trimester, commonly used for uncomplicated cystitis.
Ampicillin: Can be used but has high resistance rates among common uropathogens like E. coli. Ceftriaxone: Typically used for severe infections like pyelonephritis and administered intravenously, not necessary here. Ciprofloxacin: A fluoroquinolone, contraindicated in pregnancy due to potential cartilage damage to the fetus. Doxycycline: A tetracycline, contraindicated in pregnancy due to risks of fetal bone growth inhibition and teeth discoloration.

\vspace{0.1cm}
\textbf{Step 4: Select the best treatment} Nitrofurantoin is the safest and most effective option for this patient in the second trimester.

\vspace{0.1cm}
\textbf{Step 5: Conclusion} The best treatment for this patient is nitrofurantoin.
The answer is E: Nitrofurantoin.

\end{tcolorbox}

\caption{An example of a medical question that requires multi-step reasoning to answer.} 
\label{fig:motivation}

\end{figure*}

\subsection{Prompt of Different Actions in MCTS}
\label{subsec:a7}

\begin{figure*}[ht]
\begin{tcolorbox}[colback=white, colframe=black, width=\textwidth, title=Action 1. Propose an one-step thought.]
\#\#\# Instruction:
A 46-year-old woman comes to the physician because of a 2-week history of diplopia and ocular pain when reading the newspaper. She also has a 3-month history of amenorrhea, hot flashes, and increased sweating. She reports that she has been overweight all her adult life and is happy to have lost 6.8 kg (15 lb) of weight in the past 2 months. Her pulse is 110/min, and blood pressure is 148/98 mm Hg. Physical examination shows moist palms and a nontender thyroid gland that is enlarged to two times its normal size. Ophthalmologic examination shows prominence of the globes of the eyes, bilateral lid retraction, conjunctival injection, and an inability to converge the eyes. There is no pain on movement of the extraocular muscles. Visual acuity is 20/20 bilaterally. Neurologic examination shows a fine resting tremor of the hands. Deep tendon reflexes are 3+ with a shortened relaxation phase. Which of the following is the most likely cause of this patient's ocular complaints? A: Granulomatous inflammation of the cavernous sinus, B: Abnormal communication between the cavernous sinus and the internal carotid artery, C: Glycosaminoglycan accumulation in the orbit, D: Bacterial infection of the orbital contents, E: Sympathetic hyperactivity of levator palpebrae superioris

\#\#\# Response:
Let's think step by step. \\
Step 1: Identify key clinical features: A 46-year-old woman presents with diplopia, ocular pain, weight loss, amenorrhea, and hyperthyroid signs like tachycardia, enlarged thyroid gland, and hyperreflexia. \\
Step 2: Recognize underlying condition: These findings suggest Graves' disease with thyroid-associated orbitopathy (TAO). \\
Step 3: Determine cause of ocular symptoms: Prominent globes, lid retraction, and conjunctival injection are caused by immune-mediated glycosaminoglycan accumulation in the orbit, characteristic of TAO. \\
Step 4: Eliminate other options: Conditions like cavernous sinus inflammation, carotid-cavernous fistula, and bacterial infection do not fit the patient’s presentation. \\
Step 5: The most likely cause of the ocular complaints is glycosaminoglycan accumulation in the orbit. \\
Step 6: Conclusion: The answer is C: Glycosaminoglycan accumulation in the orbit.

\#\#\# Instruction: \{user question\} \\
\#\#\# Response: Let’s think step by step.

\end{tcolorbox}
\label{fig:a1}

\end{figure*}

\begin{figure*}[ht]
\begin{tcolorbox}[colback=white, colframe=black, width=\textwidth, title=Action 2: Propose the remaining thought steps / Action 4: Answer the sub-question again.]
\#\#\# Instruction:
A 1-year-old boy presents to the emergency department with weakness and a change in his behavior. His parents state that they first noticed the change in his behavior this morning and it has been getting worse. They noticed the patient was initially weak in his upper body and arms, but now he won’t move his legs with as much strength or vigor as he used to. Physical exam is notable for bilateral ptosis with a sluggish pupillary response, a very weak sucking and gag reflex, and shallow respirations. The patient is currently drooling and his diaper is dry. The parents state he has not had a bowel movement in over 1 day. Which of the following is the pathophysiology of this patient’s condition? A: Antibodies against postsynaptic nicotinic cholinergic ion channels, B: Autoantibodies against the presynaptic voltage-gated calcium channels, C: Autoimmune demyelination of peripheral nerves, D: Blockade of presynaptic acetylcholine release at the neuromuscular junction, E: Lower motor neuron destruction in the anterior horn \\

\#\#\# Response:
Let's think step by step. The infant has rapidly progressing weakness starting from the upper body and moving downward, bilateral ptosis, sluggish pupils, weak sucking and gag reflexes, shallow respirations, drooling, constipation (dry diaper and no bowel movement), and decreased muscle strength. These symptoms are consistent with infant botulism, caused by ingestion of Clostridium botulinum spores (often from honey) that produce botulinum toxin. The toxin inhibits the release of acetylcholine at the presynaptic terminals of the neuromuscular junction by cleaving SNARE proteins necessary for vesicle fusion. This blockade leads to flaccid paralysis and autonomic dysfunction. Therefore, the pathophysiology is blockade of presynaptic acetylcholine release at the neuromuscular junction. The answer is D: Blockade of presynaptic acetylcholine release at the neuromuscular junction. \\

\#\#\# Instruction: \{user question\} \\
\#\#\# Response: 

\end{tcolorbox}
\label{fig:a2}

\end{figure*}

\begin{figure*}[ht]
\begin{tcolorbox}[colback=white, colframe=black, width=\textwidth, title=Action 3: Propose next sub-question along with its answer.]
Given a question, please decompose it into sub-questions. For each sub-question, please answer it in a complete sentence, ending with "The answer is ". When the original question is answerable, please start the subquestion with "Now we can answer the question: <original question>".

Question 1: A 23-year-old pregnant woman at 22 weeks gestation presents with burning upon urination. She states it started 1 day ago and has been worsening despite drinking more water and taking cranberry extract. She otherwise feels well and is followed by a doctor for her pregnancy. Her temperature is 97.7°F (36.5°C), blood pressure is 122/77 mmHg, pulse is 80/min, respirations are 19/min, and oxygen saturation is 98\% on room air. Physical exam is notable for an absence of costovertebral angle tenderness and a gravid uterus. Which of the following is the best treatment for this patient? A: Ampicillin, B: Ceftriaxone, C: Ciprofloxacin, D: Doxycycline, E: Nitrofurantoin \\
Question 1.1: What is the most likely diagnosis for this patient? \\
Answer 1.1: The patient is a 23-year-old pregnant woman at 22 weeks gestation presenting with burning upon urination, known as dysuria. She has no fever, normal vital signs, and no costovertebral angle tenderness. This suggests that she is experiencing an acute uncomplicated urinary tract infection (UTI), specifically cystitis. \\
Question 1.2: Which antibiotics are safe and effective for treating cystitis during pregnancy? \\
Answer 1.2: During pregnancy, certain antibiotics are preferred due to their safety profiles for both the mother and the fetus. Nitrofurantoin is commonly used and considered safe during the second trimester for treating cystitis. Ampicillin may be less effective due to resistance. Ceftriaxone is typically reserved for more severe infections like pyelonephritis and is administered intravenously. Ciprofloxacin and doxycycline are contraindicated in pregnancy because they can harm fetal development. \\
Question 1.3: Now we can answer the question: Which of the following is the best treatment for this patient? \\
Answer 1.3: The best treatment for this pregnant patient with acute cystitis is Nitrofurantoin, as it is safe during the second trimester and effective against common urinary pathogens. Therefore, the answer is E: Nitrofurantoin. \\

Question 2: \{user question\} \\

\end{tcolorbox}
\label{fig:a3}

\end{figure*}

\begin{figure*}[ht]
\begin{tcolorbox}[colback=white, colframe=black, width=\textwidth, title=Action 5: Rephrase the question/sub-question.]
You are an AI assistant to help me rephrase questions by splitting the question context into conditions. In your rephrased question, remember to fully express the information in the original question.

Original Question: A 62-year-old woman comes to the physician because of coughing and fatigue during the past 2 years. In the morning, the cough is productive of white phlegm. She becomes short of breath walking up a flight of stairs. She has hypertension and hyperlipidemia. She has recently retired from working as a nurse at a homeless shelter. She has smoked 1 pack of cigarettes daily for 40 years. Current medications include ramipril and fenofibrate. Her temperature is 36.5°C (97.7°F), respirations are 24/min, pulse is 85/min, and blood pressure is 140/90 mm Hg. Scattered wheezing and rhonchi are heard throughout both lung fields. There are no murmurs, rubs, or gallops but heart sounds are distant. Which of the following is the most likely underlying cause of this patient's symptoms? A: Chronic decrease in pulmonary compliance, B: Local accumulation of kinins, C: Mycobacterial invasion of pulmonary parenchyma, D: Progressive obstruction of expiratory airflow, E: Incremental loss of functional residual capacity \\
Rephrased Question: A 62-year-old woman comes to the physician due to a 2-year history of coughing and fatigue. She has a morning cough productive of white phlegm and becomes short of breath when climbing stairs. She has hypertension and hyperlipidemia and recently retired from nursing at a homeless shelter. She has a 40-pack-year smoking history. Her medications include ramipril and fenofibrate. Vital signs are temperature 36.5°C (97.7°F), respirations 24/min, pulse 85/min, and blood pressure 140/90 mmHg. Physical examination reveals scattered wheezing and rhonchi throughout both lung fields. Heart sounds are distant with no murmurs, rubs, or gallops. Which of the following is the most likely underlying cause of this patient's symptoms? A: Chronic decrease in pulmonary compliance, B: Local accumulation of kinins, C: Mycobacterial invasion of pulmonary parenchyma, D: Progressive obstruction of expiratory airflow, E: Incremental loss of functional residual capacity \\

Original question: \{user question\} \\
Rephrased Question:

\end{tcolorbox}
\label{fig:a5}

\end{figure*}

\begin{figure*}[ht]
\begin{tcolorbox}[colback=white, colframe=black, width=\textwidth, title=Action 6: Generate multiple search queries and retrieve information.]
Given a question, generate a search query that would help gather information to answer it. Your goal is to formulate a query that will retrieve useful evidence or additional details that contribute to answering the question. The query should aim to obtain new information and be specific enough to ensure that the search results are relevant and helpful.

Question 1: A 23-year-old pregnant woman at 22 weeks gestation presents with burning upon urination. She states it started 1 day ago and has been worsening despite drinking more water and taking cranberry extract. She otherwise feels well and is followed by a doctor for her pregnancy. Her temperature is 97.7°F (36.5°C), blood pressure is 122/77 mmHg, pulse is 80/min, respirations are 19/min, and oxygen saturation is 98\% on room air. Physical exam is notable for an absence of costovertebral angle tenderness and a gravid uterus. Which of the following is the best treatment for this patient? A: Ampicillin, B: Ceftriaxone, C: Ciprofloxacin, D: Doxycycline, E: Nitrofurantoin \\
Query 1.1: Common causes and management of dysuria in pregnant women without systemic symptoms. \\
Document 1.1: Dysuria in pregnant women without systemic symptoms like fever or flank pain is often indicative of acute cystitis, a lower urinary tract infection typically caused by bacteria such as Escherichia coli. Pregnancy increases the risk of urinary tract infections due to hormonal changes that relax ureteral smooth muscle and mechanical compression from the enlarging uterus, leading to urinary stasis. Management involves confirming the diagnosis with a urine culture and initiating antibiotic therapy that is safe for use during pregnancy. \\
Query 1.2: Antibiotics that are safe and effective for treating cystitis during pregnancy. \\
Document 1.2: Safe and effective antibiotics for treating cystitis in pregnancy include nitrofurantoin and certain beta-lactam antibiotics like ampicillin and cephalexin. Nitrofurantoin is commonly used during the second trimester and is effective against common urinary pathogens. Ampicillin can be used but may have increased resistance rates. Ceftriaxone is a third-generation cephalosporin administered intravenously and is typically reserved for more severe infections like pyelonephritis. Antibiotics such as doxycycline (a tetracycline) and ciprofloxacin (a fluoroquinolone) are contraindicated in pregnancy due to potential risks to fetal development. \\
Query 1.3: Antibiotics contraindicated during pregnancy and their associated risks. \\
Document 1.3: Certain antibiotics are contraindicated during pregnancy due to teratogenic effects or adverse outcomes for the fetus. Doxycycline, a tetracycline antibiotic, is contraindicated because it can inhibit bone growth and cause permanent tooth discoloration in the developing fetus. Ciprofloxacin, a fluoroquinolone, is avoided as it has been associated with cartilage damage in animal studies and potential musculoskeletal risks in humans. These medications should not be used to treat infections in pregnant patients. \\

Question 2: \{user question\} \\

\end{tcolorbox}
\label{fig:a6prompt}

\end{figure*}

\begin{figure*}[ht]
\begin{tcolorbox}[colback=white, colframe=black, width=\textwidth, title=Action 7: Re-answer the question/sub-question with retrieved information.]
\#\#\# Instruction:
A 1-year-old boy presents to the emergency department with weakness and a change in his behavior. His parents state that they first noticed the change in his behavior this morning and it has been getting worse. They noticed the patient was initially weak in his upper body and arms, but now he won't move his legs with as much strength or vigor as he used to. Physical exam is notable for bilateral ptosis with a sluggish pupillary response, a very weak sucking and gag reflex, and shallow respirations. The patient is currently drooling and his diaper is dry. The parents state he has not had a bowel movement in over 1 day. Which of the following is the pathophysiology of this patient’s condition? A: Antibodies against postsynaptic nicotinic cholinergic ion channels, B: Autoantibodies against the presynaptic voltage-gated calcium channels, C: Autoimmune demyelination of peripheral nerves, D: Blockade of presynaptic acetylcholine release at the neuromuscular junction, E: Lower motor neuron destruction in the anterior horn \\

\#\#\# Relevant Documents: \\
Infant Botulism: Caused by ingestion of Clostridium botulinum spores (common sources include honey and soil). Spores germinate in the infant's gut, producing botulinum toxin. \\
Botulinum Toxin Mechanism: Blocks the release of acetylcholine at presynaptic neuromuscular junctions by cleaving SNARE proteins. Results in flaccid paralysis and autonomic dysfunction. \\
Clinical Features: Symmetric descending paralysis starting with cranial nerves (ptosis, poor suck, weak gag reflex). \\
Autonomic symptoms: constipation, dry mouth, drooling. Hypotonia and respiratory difficulties. \\

\#\#\# Response:
Let's think step by step. The infant presents with rapidly progressing weakness starting in the upper body and moving downward, bilateral ptosis with sluggish pupillary response, weak sucking and gag reflexes, shallow respirations, drooling, and constipation (dry diaper, no bowel movement). These symptoms are characteristic of infant botulism, caused by ingestion of Clostridium botulinum spores that produce botulinum toxin. The toxin inhibits the release of acetylcholine at the presynaptic terminals of the neuromuscular junction by cleaving SNARE proteins necessary for neurotransmitter release. This blockade leads to flaccid paralysis and autonomic dysfunction. Therefore, the pathophysiology is blockade of presynaptic acetylcholine release at the neuromuscular junction. The answer is D: Blockade of presynaptic acetylcholine release at the neuromuscular junction. \\

\#\#\# Instruction: \{user question\} \\
\#\#\# Relevant Documents: \{relevant documents\} \\
\#\#\# Response:

\end{tcolorbox}
\label{fig:a7prompt}

\end{figure*}

\end{document}